\documentclass[]{article}
\usepackage{amsmath}
\usepackage{amssymb}
\usepackage[T1]{fontenc}
\usepackage[utf8]{inputenc}
\usepackage{graphicx}
\usepackage{natbib}
\usepackage{color}
\usepackage{float}
\usepackage{tikz}
\usepackage{tcolorbox}
\usepackage{booktabs}
\usepackage{enumerate}
\usepackage{amsthm}
\usepackage[affil-it]{authblk}
\usepackage[onehalfspacing]{setspace}
\usepackage{hyperref}
\usepackage{geometry}
\newtheorem{theorem}{Theorem}
\newtheorem{lemma}{Lemma}
\newtheorem{proposition}{Proposition}
\newtheorem{corollary}{Corollary}
\newtheorem{definition}{Definition}

\usepackage{tikz}
\usetikzlibrary{arrows.meta, positioning, calc}

\begin{document}

\title{From Type to Token: Petri Nets, Individual Token Philosophy,
and Individualized Causal Effects}

\author[1,2]{Fernando Delbianco\thanks{Corresponding author:
\texttt{fernando.delbianco@uns.edu.ar}; ORCID: 0000-0002-1560-2587}}
\author[1,2]{Fernando Tohm\'e\thanks{ORCID: 0000-0003-2988-4519}}
\author[3,4]{Gustavo Bodanza\thanks{ORCID: 0000-0001-6254-7683}}

\affil[1]{Departamento de Econom\'ia, Universidad Nacional del Sur, Bah\'ia Blanca, Buenos Aires, Argentina}
\affil[2]{Instituto de Matem\'atica de Bah\'ia Blanca - CONICET, Bah\'ia Blanca, Buenos Aires, Argentina}
\affil[3]{Departamento de Humanidades, Universidad Nacional del Sur, Bah\'ia Blanca, Buenos Aires, Argentina}
\affil[4]{Instituto de Investigaciones Econ\'omicas y Sociales - CONICET, Bah\'ia Blanca, Buenos Aires, Argentina}

\date{}

\maketitle

\begin{abstract}
Structural Causal Models and Bayesian Networks are the dominant
framework for representing causal relations. They are fundamentally
{\em type-level} formalisms since their estimands, Average Treatment Effects and
Conditional Average Treatment Effects, are properties of distributions over
populations, not of individual units. When treatment effects exhibit unbounded
heterogeneity, these estimands collapse. What persists is the structural, {\em token-level} core of causation.

We argue that Petri Nets under the \emph{Individual Token Philosophy} are the natural representation of these token-level causal structures. We develop a formal translation from Bayesian Networks to Petri Nets, providing a two-layer account of individualized causality. The PN layer provides a structural representation of causal mechanisms that does not rest on distributional assumptions. In turn, a conformal prediction layer quantifies causal effects with finite-sample coverage guarantees. Together they constitute a complete account of individual causal effects.
\end{abstract}


\newpage

\section{Introduction}
\label{sec:intro}

What does it mean for an event to be the cause of another {\em specific} event, rather than of that {\em type} of event? The philosophy of causation has long
distinguished \emph{type causation}, as in ``smoking causes cancer'', from
\emph{token causation}, like in ``Alice's smoking caused her cancer''. The former is
a claim about regularities across populations. The latter is a claim about a
specific, concrete event. 

Pearl's framework of Structural Causal Models (SCMs) and Bayesian Networks
(BNs) is the most influential formal approach to causation in both philosophy
and the empirical sciences \citep{pearl2009,peters2017}. It provides a unified theory of
causal structure (represented by directed acyclic graphs), causal intervention (using the
$do(\cdot)$ operator), and causal identification (through the $do$-calculus). Yet
this unification comes at a cost. The SCM framework is fundamentally
type-level in its inferential output. Its estimands, the Average Treatment
Effect (ATE) and the Conditional Average Treatment Effect (CATE), are
expectations over distributions, hence properties of populations, not of
individuals.

The SCM framework \emph{can} express individual-level quantities. A unit is represented by a realization $U = u$ of
the exogenous variables, and the unit-level effect $Y_x(u) - Y_{x'}(u)$ is a genuinely token-level object \citep{pearl2009,halpern2005}. We claim that the machinery of the framework, its estimands, its identification results, and its inferential practice, is organized around distributions, and the individual appears in it only as an anonymous realization to be averaged over. The difference we press is one of emphasis on operational semantics, and the formal results below (Theorems~\ref{thm:dsep} and
\ref{thm:do}) confirm that the two frameworks agree on their subjects.

The tension between these two approaches surfaces with particular force when treatment effects exhibit
\emph{unbounded heterogeneity}. When
$\text{Var}(Y \mid do(T=t), X=x) \to \infty$, the conditional expectation
$\mathbb{E}[Y \mid do(T=t), X=x]$ no longer represents any actual individual in the population. The type-level estimand collapses. What remains is the Individual Causal Effect (ICE) $\tau_q = Y_q(1) - Y_q(0)$, a counterfactual quantity defined entirely at the token level, for the specific individual $q$.

This paper argues that Petri Nets (PNs) under the \emph{Individual Token
Philosophy} (ITP) \citep{vanglabbeek2021} are the natural formalism for this token-level causal structure. The argument proceeds in four steps. First (\S\ref{sec:typetoken}), we develop the philosophical diagnosis: SCMs contain a token-level primitive (the $do(x)$ operator) but wrap it in
probability, yielding type-level conclusions. 

In \S\ref{sec:translation}, we construct the formal translation
$T(\text{BN})$ from BNs to PNs, assuming the monotonicity of activation patterns. We show how $d$-separation maps to
flow-separation in the translated net, how $do(x)$ maps to a transition-enabling operation, and how the individual $q$ is represented by a concrete marking $M_q$ under ITP.

In Section \S\ref{sec:individualization}, we develop the individualization
argument. The ICE $\tau_q$ is the difference between two reachable markings from $M_q$ under alternative interventions, and the set of valid controls for estimating $\tau_q$ is formally characterized by equality of local subnets.

In Section \S\ref{sec:formal}, we prove the main formal results,  correspondence theorems between BN semantics and PN semantics, and a structural foundation for ensuring coverage, using conformal guarantees \citep{vovk2005}. We conclude in \S\ref{sec:discussion} by situating the framework in relation to Causal Kripke Models \citep{wang2025}, the literature on singular causation \citep{baetu2025,hall2004}, and the internal ITP/CTP debate in PN theory \citep{vanglabbeek2021,goltz1994}.

\section{From Type to Token Causation}
\label{sec:typetoken}

\subsection{The type/token distinction}

The philosophy of causation has recognised two fundamentally different kinds of
causal claims since Lewis \citeyearpar{lewis1973}. \emph{Type causation} is a
claim about kinds of events, typically analyzed in probabilistic or nomological
terms. The relata are event types, and the claim is verified by statistical
regularities across populations. \emph{Token causation} (also called singular
causation) is a claim about specific, individual events, typically analyzed via
counterfactuals. In this case, the relata are concrete event tokens and the claim is
verified, if at all, by what would have happened to \emph{this} event under
different antecedent conditions.

Hall \citeyearpar{hall2004} influentially argues that `causation' names two
distinct concepts: \emph{dependence} (counterfactual, token-level) and
\emph{production} (mechanistic, process-level). This matters for our purposes
because, as we argue, Pearl's SCM framework contains both concepts but does not
formally separate them, and this conflation is precisely what fails under
unbounded heterogeneity.

The connection between formal causal models and the type/token distinction has
received increasing attention. \citet{erasmus2023} shows that Causal Bayes Nets
(CBNs) are abstract enough to fit any token-level theory of causation, but only
under certain modelling assumptions that are not always satisfied. When those
assumptions fail, accounts of actual causation based on CBNs face fundamental
limitations. Our contribution is to formally specify those assumptions 
and to identify the formalism that obtains when they are dropped.

\subsection{The SCM as an ambiguous framework}

Pearl's SCM framework has three layers that operate at different levels. The structural equations $Y = f(X, \varepsilon)$ are type-level. They describe
how variables relate in general, for the entire population. The function $f$ is
the same for all individuals. The $do(x)$ operator is operationally token-level:
`I intervene and set $X$ to $x$ in this specific system.' The physical
interpretation is of a concrete manipulation of a particular system. But the
output of $do(x)$, the interventional distribution $P(Y \mid do(X=x))$, is
immediately type-level again, an expectation over the noise $\varepsilon$
averaging over all individuals who share the same structural equations.

The tension arises because $do(x)$ uses a token-level primitive to produce type-level
conclusions, skipping exactly the quantity that is often the object of interest.
This is not to say that the framework cannot recover that quantity: fixing the
exogenous realization $U = u$ yields the unit-level effect, as noted in
\S\ref{sec:intro}. The point is that this move runs against the grain of the
formalism, whose semantics and standard theorems are stated for distributions,
whereas in the process semantics developed below the individual is the default
object and the distribution is the derived one.
\citet{baetu2025} argues that
knowledge of singular causation depends epistemically on type-level
population contrasts. We accept the epistemic dependence while denying the
ontological one, since $\tau_q = Y_q(1) - Y_q(0)$ is a property of an individual $q$,
not of a population distribution. The population serves as epistemic scaffolding without being
a {\em constituent} of the causal fact.

The modal-relativist analysis of \citet{hitchcock2024} adds further texture. What
counts as the correct causal model depends on which background possibilities are
held fixed, a specification that is context-dependent and not fixed by the
model itself. The `local SCM' of individual $q$, which can be formalized in our framework of Section \S\ref{sec:translation},
as a subnetwork of the translated PN.

\subsection{The empirical trajectory: from type to token by necessity}

Assessing the answer to a specific query $q$ using the full population is suboptimal, only the subset of observations that
share the local mechanism of $q$ are relevant \citep{delbianco2021}. This implicitly recognizes that
$q$ has its own mechanism, distinct from the population mechanism. In the causal setting, this means that the global SCM does not apply to the individual $q$.

In \citet{delbianco2023}, this is formalized via conformal prediction. $q$
defines its own proxy population by structural similarity, and inference is
calibrated on that local class. The conformal guarantee is distribution-free but
requires exchangeability within the local class. This can be further developed, formalizing local conformal prediction for individual causal effects under distribution shift, precisely the estimation problem that the quantitative layer of the present framework, introduced in \S\ref{sec:individualization}, is designed to support.

A limit is reached when
$\text{Var}(Y \mid do(T=t), X=x) \to \infty$. Then, the Conditional ATE collapses. The
conditional expectation no longer represents any individual. The only meaningful
causal quantity that survives is $\tau_q = Y_q(1) - Y_q(0)$, a purely
token-level quantity. The SCM retains its structural role, but its probabilistic
layer ceases to be informative.

\subsection{Two token philosophies in Petri Nets: an internal mirror}

Remarkably, Petri net theory has independently arrived at the same fundamental
distinction. \citet{vanglabbeek2021} identify two competing philosophies for
the causal semantics of PNs:

\begin{itemize}

\item The \emph{Individual Token Philosophy} (ITP): it matters which specific token
participates in each transition. If token $A$ fires transition $T_1$ and token
$B$ fires $T_2$, their causal histories will be distinct even if they originate from
the same place. Causal dependence is tracked at the level of individual tokens.

\item The \emph{Collective Token Philosophy} (CTP): only the number of tokens in each
place matters, not which ones. This abstracts away individual identity. This is analogous to type causation since what matters is that \emph{some} token of the right
type is present.
\end{itemize}

When heterogeneity is unbounded, this
CTP-analogue fails. The ICE requires ITP. The trajectory of  the \emph{specific}
token ($q$) is what matters.

ITP was introduced to distinguish causal semantics of
Petri nets \citep{vanglabbeek2021}. What ITP supplies is
exactly the formal resource that token causation demands, persistent identity
and traceable causal history for the entities the formalism reasons about.
Whether that resource can be made to represent \emph{individuals}, rather than
merely tokens, requires a bridge principle, which we state explicitly in
\S\ref{sec:translation} when the individual $q$ is introduced as a marking.

\section{Petri Nets as a Token-Level Causal Formalism}
\label{sec:translation}

\subsection{Basic formalism}~\label{subsec: basic}

A Petri Net is a tuple $(P, T, F, M_0)$ where $P$ is a finite set of
\emph{places}, $T$ is a finite set of \emph{transitions} (with $P \cap T =
\emptyset$), $F \subseteq (P \times T) \cup (T \times P)$ is the \emph{flow
relation}, and $M_0 : P \to \mathbb{N}$ is the \emph{initial marking}, assigning
to each place a number of tokens.

A transition $t \in T$ is \emph{enabled} under a marking $M$ if every input
place has at least one token, i.e., for all $p$ such that $(p,t) \in F$, $M(p) \geq 1$.
When $t$ fires, it consumes one token from each input place and produces one
token in each output place, yielding a new marking $M'$. We write $M \xrightarrow{t} M'$.
The set of markings \emph{reachable} from $M_0$ is denoted $R(M_0)$. Note that this firing rule is conjunctive: a transition with multiple output arcs deposits a token in every output place. Exclusive branching between alternative outcomes is represented not by a single transition with multiple outputs, but by conflict, distinct transitions competing for the same input token, as is standard in free-choice nets \citep{vanglabbeek2021}.

Under the ITP, tokens have identity. Two tokens in the same place are
distinguishable by their \emph{causal history}, the sequence of transitions
whose firings produced them. The marking $M_q$ under ITP records which individual tokens, with their corresponding histories, occupy each place. 

\subsection{The translation $T(\text{BN})$}

A Bayesian Network (BN) is a directed acyclic graph $G=(V,E)$ together with a joint probability distribution $P$ over the variables $V$, where each node represents a random variable and each directed edge encodes a direct probabilistic dependence. Under the causal Markov condition, every variable is conditionally independent of its non-descendants given its parents, allowing the joint distribution to factorize as
\[
P(V)=\prod_{i=1}^{n} P(X_i \mid \mathrm{pa}(X_i)),
\]
where $\mathrm{pa}(X_i)$ denotes the set of parents of $X_i$ in the graph.

Given a BN $= (V, E, \mathcal{P})$ with binary variables $V = \{X_1, \ldots, X_n\}$
under the Markov condition, we define a translation $T(\text{BN})$ satisfying the following condition:

\begin{definition}[Monotone support condition]~\label{def:msc} For $X_i$, let
	\[
	\mathcal{C}_i^{+}
	=
	\left\{
	c\in\{0,1\}^{|\operatorname{pa}(X_i)|} :
	P(X_i=1\mid \operatorname{pa}(X_i)=c)>0
	\right\}.
	\]
    
\noindent such that
\[c\in\mathcal{C}_i^{+} \ \mbox{and} \ 
		c\leq c' \ \mbox{componentwise} \ \Rightarrow  c'\in\mathcal{C}_i^{+}\]
\noindent while		
		\[
		(0,\ldots,0)\notin\mathcal{C}_i^{+}.
		\]
	\end{definition}

Then, the components of $T(\text{BN}) = (P_{\text{PN}}, T_{\text{PN}}, F, M_0)$ are defined as follows:

\begin{itemize}

\item \textbf{Places}: For each node $X_i \in V$, a place $p_i \in P_{\text{PN}}$.
A token in $p_i$ represents $X_i = 1$. Its absence represents $X_i = 0$.

\item \textbf{Transitions}: For each variable $X_i$ with parents
$\text{pa}(X_i) = \{X_{j_1}, \ldots, X_{j_k}\}$ and each minimal element $c \in C_{i}^{+}$ we create a transition
$t_{i,c}$.

\item \textbf{Arcs}: For the transition $t_{i,c}$:
\begin{itemize}
\item An input arc from $p_{j_l}$ whenever $c_l = 1$.
  \item An output arc to $p_i$
\end{itemize}

\item \textbf{Initial marking}: $M_0(p_i) = 1$ if $X_i$ is an active exogenous root
variable, 0 otherwise.
\end{itemize}

In $T(\text{BN})$, arcs from parent places to transitions are interpreted as {\em persistent}. That is, the presence of a parent node is tested without consuming its token. Otherwise a parent node with two children may not be able to activate both of them, spending a token for just one of them. This extends the PNs defined in ~\ref{subsec: basic} to what are known as {\em contextual nets} \citep{Montanari1995}.\footnote{Because arcs from parent places are non-consuming, two distinct minimal configurations $t_{i,c_1}$ and $t_{i,c_2}$ for the same $X_i$ may in principle both become enabled and
fire, each depositing a token in $p_i$. This does not affect any result below, since Theorems 1 and 2 and Propositions 1–2 only test $M(p)\geq 1$, never an exact count. We nonetheless treat places as 1-bounded throughout,
consistent with their Boolean reading in Definition~1: once $p_i$ carries a token, later firings of other transitions targeting $p_i$ are inert for the purposes of the marking's interpretation.}  Each transition $t_{i,c}$ encodes a \emph{minimal causally sufficient configuration}, represented by the combination of active parents that {\em may} produce $X_i = 1$. The configurations supported in a BN are translated into reachable markings. In turn, individual firing sequences record their token-level causal realization.

$T(\text{BN})$ is a \emph{structural} translation. The Markov
condition is used to construct the net, since it licenses reading each
conditional $P(X_i \mid \mathrm{pa}(X_i))$ as a local mechanism, and it is the
resulting mechanisms. No independencies, factorizations, or probabilities appear in the
translated net. The Markov
condition survives in qualitative form, as the correspondence between
$d$-separation and flow-separation that will be established in Theorem~\ref{thm:dsep} and that
Corollary~\ref{cor:soundness} makes precise, while the probabilistic layer is
reintroduced, where quantification demands it, by the conformal layer of
\S\ref{sec:individualization}. 

If the only payoff of $T(\text{BN})$ were ``now we have tokens'', the
translation would amount to a change of notation, since any sufficiently rich
formalism can encode a DAG. The payoff lies in properties that are key in
PNs and not quite central in
the BN formalism, yielding an \emph{operational semantics}, under
which mechanisms are executable and a causal history is a concrete run of the
net. It also provides \emph{token identity}, which grounds the representation of the individual
$q$ in \S 3.4 below; \emph{concurrency}, which distinguishes
mechanisms that fire independently from those that merely happen to be
uncorrelated; and the possibility of \emph{cycles}, which BNs exclude by
construction (\S\ref{sec:discussion}). Each of these is put to work in what
follows, and we return to their joint significance in
\S\ref{sec:discussion}.

A different line of work in theoretical computer science relates Petri nets and
Bayesian networks in the opposite direction. \citet{bruni2020} equip Petri
occurrence nets with probability distributions and use Bayesian-network
inference machinery to resolve \emph{confusion}, the interplay of conflict and
concurrency that confuses non-deterministic choice. This is solved via a
decomposition of the net into structural branching cells. Their target is a
compositional probabilistic semantics for concurrent systems in general. The
correspondence they establish is with the algorithms for eliminating
nondeterminism. The translation
$T(\text{BN})$ developed here instead starts from a given causal BN and asks
what token-level structure preserves its causal semantics, so that
individual-level counterfactuals become representable. The two projects are
compatible rather than competing. Their apparatus for resolving confusion
could in principle be combined with $T(\text{BN})$ to give a fully
probabilistic treatment of concurrent choice within the translated net, a
direction we leave for future work.

\subsection{The $do(x)$ operation in the PN}

In Pearl's framework, $do(X_k = 1)$ over a BN consists of: (i) removing all
incoming edges to node $X_k$, and (ii) fixing $X_k = 1$. The PN translation is defined as follows:

\begin{definition}[$\mathrm{do\_PN}$]
The intervened PN $\mathrm{do\_PN}(p_k = x)(M_q)$ is obtained by:
\begin{enumerate}
  \item Removing all arcs $(p, t) \in F$ and disable every transition $t$ that would determine
        $p_k$ as output (severing the mechanism that normally determines $p_k$). 
  \item Setting $M_q(p_k) = x$ and $M_q(p_j) = M_q(p_j)$ for all $j \neq k$
\end{enumerate}
\end{definition}

This definition mirrors Pearl's mutilation, and
Theorem~\ref{thm:do} certifies that the rewriting is faithful. The operation is token-level in both its operational semantics and its output. It modifies the concrete marking $M_q$ of individual $q$. The interventional distribution $P(Y \mid do(X_k = x))$ in the
BN is the type-level abstraction of this token-level operation, obtained by marginalizing over noise realizations. 

\subsection{The individual $q$ as a marking}

Under ITP, the individual query $q$ with observed feature vector $x_q$ has a
natural formal representation as a concrete marking $M_q$ of $T(\text{BN})$:
\[
  M_q(p_i) =
  \begin{cases}
    1 & \text{if } X_i = 1 \text{ for individual } q \\
    0 & \text{if } X_i = 0 \text{ for individual } q
  \end{cases}
\]
This marking is the causal state of $q$'s world, a concrete world. This concreteness is precisely what ITP makes possible and what the CTP reading of a net abstracts away. In the BN, the corresponding object is a complete instantiation of the variables, which the probabilistic semantics treats as one anonymous realization among many rather than as a first-class object of the formalism.

This representation rests on a bridge principle that deserves to be made
explicit, as announced in \S\ref{sec:typetoken}. ITP guarantees that tokens
have identity and causal history, but it does not by itself say that a token
\emph{is} an individual. We therefore adopt the following representational
postulate: \emph{the individual $q$ is represented by the marking $M_q$, and
the identity of the tokens in $M_q$ tracks the identity of $q$'s actual
features together with the causal histories that produced them}. The justification of this postulate is that it supplies exactly what the ICE
requires. $\tau_q$ is defined by re-running the \emph{same} individual under
alternative interventions, and ``the same individual'' is formally meaningful
only in a semantics where identity persists across firings. CTP cannot ground
the postulate, since it identifies markings that differ only in which tokens
occupy which places. ITP is the weakest PN semantics that can achieve this.

The two counterfactual markings that define the ICE are:
\[
  M_q^{T=1} = \mathrm{do\_PN}(p_T = 1)(M_q)
  \qquad \text{and} \qquad
  M_q^{T=0} = \mathrm{do\_PN}(p_T = 0)(M_q)
\]
The ICE is then represented structurally as:
\[
  \tau_q \;\longleftrightarrow\;
  R(M_q^{T=1})(p_Y) - R(M_q^{T=0})(p_Y)
\]
where $R(M)(p) \in \{0,1\}$ indicates whether $p$ carries a
token in the terminal marking reachable from $M$. 

Notice that when reachability is non-deterministic, the terminal marking depends on an adequate definition of $R(M)(p)$ as follows:
\[
  R(M)(p) =
  \begin{cases}
    1 & \text{if there exists } M' \in R(M) \ \text{with} \ M'(p) \geq 1, \\
    0 & \text{otherwise. } 
  \end{cases}
\]
\section{The Individualization Argument}
\label{sec:individualization}

\subsection{The ICE as a token-level quantity}

The progression ATE $\to$ CATE $\to$ ICE is often presented as statistical
refinement, from a population estimand to a conditional one, and then to an
individual one. But this reading obscures the philosophically salient fact. The
ATE and CATE are parameters of distributions (type-level objects). The ICE is
a counterfactual fact about a concrete individual (a token-level object). A distributional
parameter is estimated from population data. A counterfactual fact about $q$ is
a property of $q$, and the population serves only as epistemic scaffolding for
its estimation.

This distinction determines what kind of formalism is needed. A distributional
parameter can be represented in a probabilistic structure. A counterfactual
fact about an individual requires a formalism where that individual is a
first-class object. PNs under ITP provide exactly this. $M_q$ is the
individual, transitions are its causal mechanisms, and reachable markings are
its counterfactual worlds.

\subsection{The local SCM of $q$ as a subnetwork}

The empirical procedure of \citet{delbianco2021,delbianco2023}
constructs, for each query $q$, a set of relevant controls $\mathcal{R}_q$ that
share the local causal mechanism of $q$. In the language of
\citet{peters2016}, this is equivalent to discovering the \emph{local SCM} of
$q$, the restriction of the global SCM to the structural neighborhood of $q$.

This operation has a precise PN representation. The local SCM of $q$ is the
\emph{reachable subnetwork} from $M_q$ in $T(\text{BN})$:

\begin{definition}[Local Net]
\label{def:localnet}
\[
  \mathrm{LocalNet}(q) = \bigl(P_q,\, T_q,\, F_q\bigr)
\]
where $P_q = \{p \in P_{\text{PN}} : M_q(p) \geq 1 \text{ or }
p \in R(M_q)\}$, $T_q$ are the transitions that can be enabled in some marking
of $R(M_q)$, and $F_q = F \cap \bigl((P_q \times T_q) \cup (T_q \times P_q)\bigr)$.
\end{definition}

The relevant controls $\mathcal{R}_q^*$ are, in PN terms, the individuals whose
initial marking generates the same reachable subnetwork. Under ITP, this is a property of individual tokens. Two individuals that have the same feature values $X$ but arrived
at them via different causal trajectories may have different LocalNets and
therefore be invalid controls for one another, even if superficially similar
in covariates.

Equality of LocalNets is a strong criterion, and weaker
alternatives are conceivable: isomorphism of the local subnets rather than
identity, or agreement only up to a bounded radius around the treatment and
outcome places. We work with exact equality because it is the weakest
condition under which the invariance result of
Proposition~\ref{prop:localnet}(b) holds without further assumptions. 

\subsection{Two counterfactuals as intervened markings}

The fundamental problem of causal inference, that we cannot simultaneously observe $Y_q(1)$ and $Y_q(0)$, has a particularly clear formulation in the PN
framework.

Starting from the concrete marking $M_q$, the two interventions
$\mathrm{do\_PN}(p_T = 1)$ and $\mathrm{do\_PN}(p_T = 0)$ produce two mutilated
nets with modified markings $M_q^{T=1}$ and $M_q^{T=0}$. These are two causally
possible worlds for $q$, generated by the same mechanistic structure under
different interventions. The impossibility of observing both simultaneously is
a structural property of the PN since the same
marking $M_q$ cannot evolve in two different directions in a single run.

The conformal prediction approach \citep{vovk2005} estimates the
unobserved counterfactual by using $\mathcal{R}_q^{*(t)}$, individuals with
compatible LocalNet under intervention $do(T=t)$, as an empirical approximation
of $q$'s alternative world. The coverage guarantee of the conformal interval
$\hat{C}(\tau_q)$ rests on the exchangeability of non-conformity scores within
$\mathcal{R}_q^{*(t)}$, which in turn rests on those individuals sharing $q$'s
LocalNet.

\subsection{What the PN formalism adds: structure without distribution}

Unlike in the BN/SCM framework, where structure and distribution are entangled, in the PN, the separation is explicit. The claim `$X$ causes $Y$ through
mechanism $f$' is represented by the existence of a transition $t$ with an
input arc from $p_X$ and an output arc to $p_Y$, a structural claim, verifiable
without distributions. When outcome distributions are unknown or unbounded, this
structure remains informative. This motivates a two-layer account:

\emph{The structural layer} (PN under ITP) determines which causal mechanisms
are relevant for $q$ ($\mathrm{LocalNet}(q)$), which interventions have effect
(which transitions are enabled by $\mathrm{do\_PN}$), and which counterfactuals
are pertinent ($R(M_q^{T=1})$ vs. $R(M_q^{T=0})$). This layer requires no
probability distributions.

\emph{The quantitative layer} (conformal prediction) determines how much $q$'s
outcome changes under the alternative intervention, with coverage guarantee
$\geq 1-\alpha$. This layer is also distribution-free but uses data, the
relevant controls $\mathcal{R}_q^{*(t)}$, as empirical reference.

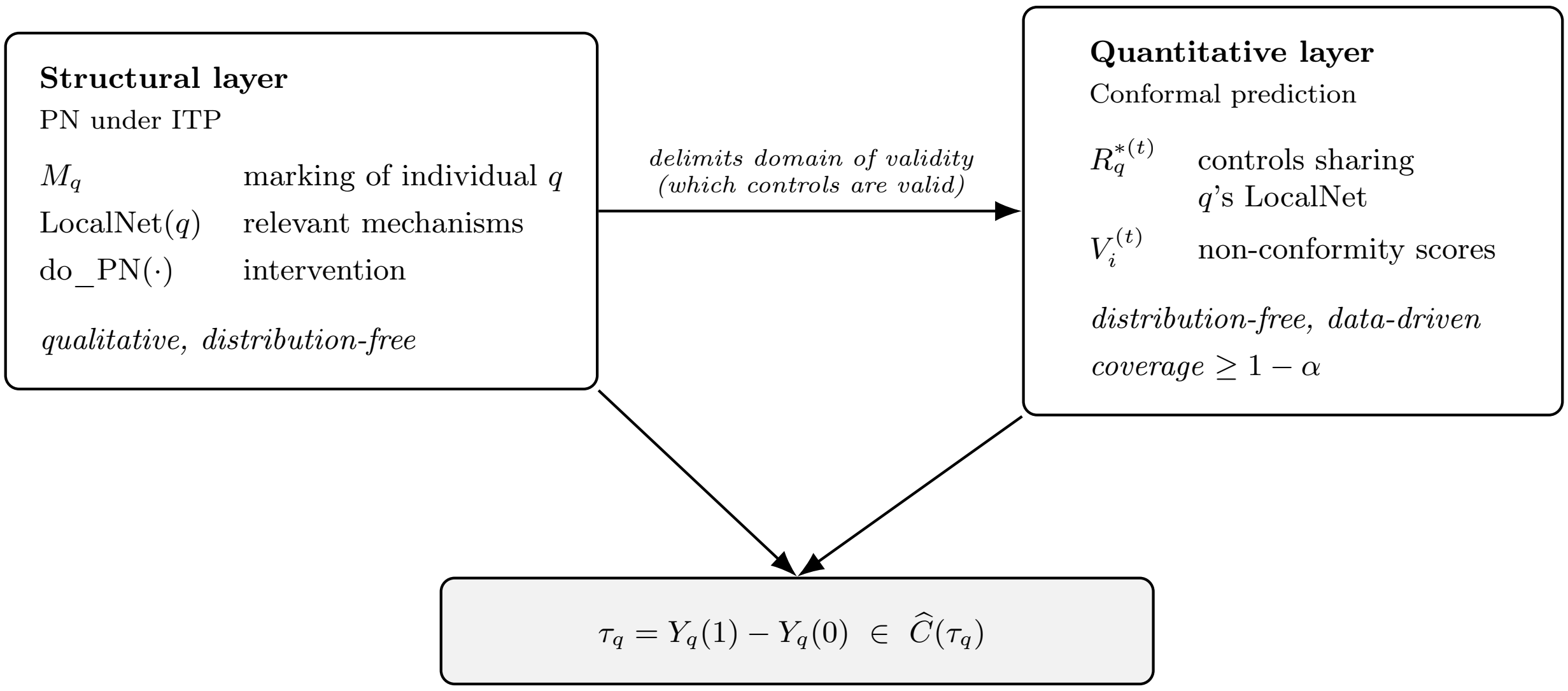
\begin{figure}[t]
\centering
\begin{tikzpicture}[
  box/.style={draw, rounded corners, align=left, minimum width=5.6cm,
              minimum height=3.6cm, inner sep=10pt, line width=0.8pt},
  result/.style={draw, rounded corners, align=center, minimum width=7.4cm,
              minimum height=1.1cm, fill=black!5, line width=0.8pt},
  arr/.style={-{Latex[length=2.8mm]}, thick},
  lbl/.style={font=\scriptsize\itshape, align=center},
  font=\small
]

\node[box] (struct) at (0,0) {
  \textbf{Structural layer} \\[1pt]
  {\footnotesize PN under ITP} \\[8pt]
  \begin{tabular}{@{}ll@{}}
    $M_q$                       & marking of individual $q$ \\[3pt]
    $\mathrm{LocalNet}(q)$      & relevant mechanisms \\[3pt]
    $\mathrm{do\_PN}(\cdot)$    & intervention \\
  \end{tabular}\\[10pt]
  {\itshape qualitative, distribution-free}
};

\node[box, right=4.4cm of struct] (quant) {
  \textbf{Quantitative layer} \\[1pt]
  {\footnotesize Conformal prediction} \\[8pt]
  \begin{tabular}{@{}ll@{}}
    $R^{*(t)}_q$   & controls sharing \\
                   & $q$'s LocalNet \\[3pt]
    $V^{(t)}_i$    & non-conformity scores \\
  \end{tabular}\\[10pt]
  {\itshape distribution-free, data-driven}\\[3pt]
  {\itshape coverage $\ge 1-\alpha$}
};

\node[result, below=1.8cm of $(struct.south)!0.5!(quant.south)$] (tau) {
  $\tau_q = Y_q(1) - Y_q(0) \ \in\ \widehat{C}(\tau_q)$
};

\draw[arr] (struct.east) -- (quant.west)
  node[midway, above, lbl, text width=3.8cm] {delimits domain of validity\\ (which controls are valid)};

\draw[arr] (struct.south east) -- (tau.north);
\draw[arr] (quant.south west) -- (tau.north);

\end{tikzpicture}
\caption{The two-layer account of the Individual Causal Effect. The structural layer (PN under ITP) determines which mechanisms and controls are relevant to $q$ without appeal to distributions; the quantitative layer (conformal prediction) uses that structural constraint to deliver a finite-sample coverage guarantee for $\tau_q$.}
\label{fig:twolayer}
\end{figure}

The two layers articulate as follows. The structural layer delimits the domain
of validity of the quantitative layer (only individuals with compatible LocalNet
are valid controls), and the quantitative layer operationalizes the counterfactual
that the structural layer defines qualitatively. This is illustrated in Figure \ref{fig:twolayer}.

A caveat guards this claim against overstatement. What the structural layer
dispenses with is distributional assumptions for \emph{representing} causal
structure. \emph{Quantifying} the magnitude of $\tau_q$ still requires data,
as the quantitative layer makes explicit. The proper contrast is therefore not
``with versus without distributions'' but between structural representation
and statistical quantification, and the contribution of the PN formalism is to
make the boundary between the two exact.

\section{Formal Results}
\label{sec:formal}

\subsection{Preliminary definitions}

\begin{definition}[Path in $T(\text{BN})$]
A \emph{path} in $T(\text{BN})$ is an alternating sequence
$\pi = p_0, t_1, p_1, t_2, \ldots, t_n, p_n$ where each consecutive pair of places
is connected by an arc in $F$. A place $p_k$ ($0 < k < n$) is a
\emph{collider node} in $\pi$ if both adjacent arcs in $\pi$ are incoming:
$(p_{k-1} \to t_k)$ and $(p_{k+1} \to t_{k+1})$, with $t_k$ and $t_{k+1}$
having output arcs to $p_k$. It is a \emph{non-collider} otherwise.
\end{definition}

\begin{definition}[Flow-active path under ITP]~\label{def:f-ap}
A path $\pi$ is \emph{flow-active} relative to a set $Z$ of places if:
\begin{enumerate}
  \item Every non-collider $p_k$ ($0 < k < n$) satisfies $p_k \notin Z$
  \item Every collider $p_k$ satisfies $p_k \in Z$, or some place in
        $R(p_k)$, the set of places reachable by firing transitions from
        $p_k$, belongs to $Z$
\end{enumerate}
Condition (2) requires ITP: under ITP, a token in a collider place has an
identifiable causal history (it arrived via one of the input transitions), so
knowing the collider's state opens information about its sources. Under CTP
this is not possible, and condition (2) cannot be implemented.
\end{definition}

\begin{definition}[Flow-separation]
Places $p_i$ and $p_j$ are \emph{flow-separated} by $Z$ in $T(\text{BN})$
under ITP if there is no flow-active path between $p_i$ and $p_j$ relative
to $Z$.
\end{definition}

\subsection{$d$-Separation and flow-separation}

\begin{definition}[Edge effectiveness]~\label{def:ee}

A BN is {\em edge-effective} if for every edge $X_j \rightarrow X_i$ there exists $c^{*} \in \min({\mathcal C}_i^{+})$ such that $c^{*}_j =1$. That is, if every edge in the BN occurs in at least one minimal sufficient configuration.
\end{definition}

Edge-effectiveness requires every graphical edge to participate in at least one minimal sufficient activation mechanism. It is weaker than {\em faithfulness}, the property that the probability distribution in BN has no conditional independencies beyond those encoded by $d$-separation in the graph \citep{Spirtes2001}, since it imposes no restrictions on conditional independencies among non-adjacent variables or on cancellations along multiple paths. It just rules out edges that are inert with respect to the positive-support semantics represented by $T(\mathrm{BN})$.

\begin{theorem}[$d$-Separation Correspondence]
\label{thm:dsep}
Let $\mathrm{BN} = (V, E, \mathcal{P})$ be a Bayesian network satisfying the
Markov condition and edge-effectiveness.  Let $T(\mathrm{BN})$ be its PN translation under ITP.
For any disjoint sets $\mathbf{X}, \mathbf{Y}, \mathbf{Z} \subseteq V$:
\[
  \mathbf{X} \perp_d \mathbf{Y} \mid \mathbf{Z} \;\text{ in BN }
  \iff
  \{p_x : X_x \in \mathbf{X}\} \text{ and } \{p_y : X_y \in \mathbf{Y}\}
  \text{ are flow-separated by } \{p_z : X_z \in \mathbf{Z}\} \text{ in } T(\mathrm{BN})
\]
\end{theorem}

To prove this result, we need a previous one:

\begin{lemma}[Reachability correspondence]
\label{lem:reach}
For any $X_i, X_k \in V$, $X_i$ is a descendant of $X_k$ in the BN graph $G$, where BN satisfies the conditions of Theorem~\ref{thm:dsep}
if and only if $p_i \in R(p_k)$ in $T(\mathrm{BN})$.
\end{lemma}
\begin{proof}
By construction, $X_k \in \mathrm{pa}(X_i)$ implies a transition $t_{i,c}$ with
an input arc from $p_k$ and output arc to $p_i$, for any configuration $c$
compatible with $X_k=1$ and $P(X_i=1\mid c)>0$. Hence $p_i \in R(p_k)$
whenever $X_i$ is a child of $X_k$. The general case follows by induction on
the length of the directed path from $X_k$ to $X_i$ in $G$.
\end{proof}

Then we have:

\begin{proof}[Proof of Theorem~\ref{thm:dsep}]
By induction on path length. The base case ($n=2$) covers the three fundamental
path types.

\emph{(i) Fork} ($p_i \leftarrow t_{ki} \leftarrow p_k \rightarrow t_{kj}
\rightarrow p_j$, with $p_k$ a non-collider). If $p_k \in \mathbf{Z}$:
condition (1) of flow-activity fails, path is blocked; in BN,
$X_i \perp X_j \mid X_k$ follows from the Markov condition applied to the
fork. If $p_k \notin \mathbf{Z}$: path is flow-active. In BN,
the fork path is open and induces dependence. 

\emph{(ii) Chain} ($p_i \rightarrow t_{ik} \rightarrow p_k \rightarrow t_{kj}
\rightarrow p_j$, with $p_k$ a non-collider). Analogous to the fork case by
the symmetry of blocking conditions. 

\emph{(iii) Collider} ($p_i \rightarrow t_i \rightarrow p_k \leftarrow t_j
\leftarrow p_j$, with $p_k$ a collider). If $p_k \notin \mathbf{Z}$ and no
descendant of $p_k$ is in $\mathbf{Z}$, condition $(2)$ of Definition~\ref{def:f-ap} fails and the path gets blocked.
In BN, the unconditioned collider blocks dependence. If $p_k \in \mathbf{Z}$:
under ITP, the token in $p_k$ has identifiable causal history, it arrived via
$t_i$ or $t_j$, so knowing $p_k$'s state opens information about $p_i$ and
$p_j$. Condition (2) is satisfied. The path is flow-active. In BN, conditioning on
the collider opens dependence. This is an instance of Berkson's paradox \citep{sharmaetal2022}. 

For the inductive step, consider an arbitrary path of length $n>2$. Each
internal place is classified, by the direction of its two incident arcs, as
a collider or non-collider. By construction of $T(\mathrm{BN})$, this
classification agrees with the corresponding node's status in the BN path.
By the induction hypothesis, the flow-activity conditions for the shorter
subpaths obtained by removing an endpoint agree with the $d$-separation
conditions. For collider places, agreement of condition (2) additionally
requires Lemma~\ref{lem:reach}, which guarantees that $R(p_k)$ in $T(\mathrm{BN})$
coincides with the descendant set of $X_k$ in $G$. Since a path is active
exactly when every non-collider lies outside $Z$ and every collider is either already in $\mathbf{Z}$  or has a descendant in $\mathbf{Z}$, the conjunction of these 
local conditions yields the equivalence for the full path.
\end{proof}

\begin{corollary}[Soundness]
\label{cor:soundness}

Under the Markov condition, flow-separation in $T(\mathrm{BN})$ under ITP implies
conditional independence in the BN distribution: if $\{p_x\}$ and $\{p_y\}$
are flow-separated by $\mathbf{Z}$, then
$P(X_i, X_j \mid \mathbf{Z}) = P(X_i \mid \mathbf{Z}) \cdot P(X_j \mid \mathbf{Z})$.
\end{corollary}

Theorem~\ref{thm:dsep} fails under CTP. The collider case requires tokens to
have identifiable causal histories, a property that CTP explicitly abstracts
away. Under CTP, conditioning on $p_k$ cannot open the information channel
between $p_i$ and $p_j$, so the correspondence breaks precisely where the
ITP/CTP distinction matters. This confirms that type-level semantics (CTP) are
formally inadequate for token-level causal reasoning.

A natural objection is that flow-activity is defined so as to mimic $d$-separation, making Theorem~\ref{thm:dsep} true by construction. But notice, first, that the clause that does the distinctive work,
condition (2) on colliders, is not an arbitrary stipulation but is forced by
the execution semantics of the net under ITP. A token in a collider place
carries an identifiable causal history, so conditioning on the collider
genuinely reveals information about its sources. The definition records a fact
about the semantics, and that the same clause is
\emph{unimplementable} under CTP shows it is not free to stipulate. Second,
the theorem should be read as an adequacy result, certifying that the token-level semantics recovers exactly the
independence structure of the type-level formalism. Its
value is the guarantee
that nothing of the BN's qualitative content is lost by the
translation.

\subsection{Correspondence for the $do(x)$ operator}

\begin{theorem}[do$(x)$ Correspondence under monotone support]
	\label{thm:do}
	Consider $\mathrm{BN}$ with binary variables
	satisfying the Markov condition. 
	Let $M_q$ be a marking representing individual $q$.  For any variable
	$X_k\in V$ and $x\in\{0,1\}$, let
	$\mathrm{BN}_{do(X_k=x)}$ denote the mutilated BN, and $do_{\mathrm{PN}}(p_k=x)(M_q)$
	denote the PN intervention.  Then, for every
	outcome variable $X_j$,
	\[
	P\!\left(
	X_j=1\mid do(X_k=x),M_q
	\right)>0
	\quad\Longleftrightarrow\quad
	\exists M'\in
	R\!\left(do_{\mathrm{PN}}(p_k=x)(M_q)\right)
	\text{ such that }M'(p_j)\geq 1.
	\]
	That is, $X_j=1$ belongs to the support of the interventional
	distribution if and only if $p_j$ can be marked in the intervened
	PN.
\end{theorem}

\begin{proof}[Proof]
Consider the mutilated DAG associated with
	$\mathrm{BN}_{do(X_k=x)}$.  Since the graph is acyclic, its variables
	can be arranged in order. 

    The proof is by induction. For root variables, the claim follows directly from the initial marking.
	The intervened variable $X_k$, is fixed to $x$ in the mutilated
	$BN$, while the marking of $p_k$ is fixed to the same value in the
	intervened $PN$.
	
	Now consider a non-root, non-intervened variable $X_i$, and assume that
	the correspondence has been established for all its predecessors. Suppose first that
	\[
	P\!\left(X_i=1\mid do(X_k=x),M_q\right)>0.
	\]
	By the factorization of the mutilated BN, there must exist a parent
	configuration $c$ having positive probability under the intervention
	such that
	\[
	P\!\left(
	X_i=1\mid \operatorname{pa}(X_i)=c
	\right)>0.
	\]
	Hence $c\in\mathcal{C}_i^{+}$.  Since
	$\mathcal{C}_i^{+}$ is finite, there exists a minimal element
	$c^{*}\in\min(\mathcal{C}_i^{+})$ satisfying 
	$c^{*}\leq c$.
	By construction of $T(\mathrm{BN})$, there is a transition
	$t_{i,c^{*}}$ whose input places are precisely the places
	$p_{j_\ell}$ for which $c^{*}_\ell=1$.
	
	Because $c$ has positive support, all the parents required by
	$c^{*}$ can occur simultaneously.  By induction, the
	corresponding places can therefore be simultaneously marked in a
	reachable marking of the intervened PN.  The transition
	$t_{i,c^{*}}$ is consequently enabled.  Its firing produces a token
	in $p_i$, and hence there exists $M'\in
	R\!\left(do_{\mathrm{PN}}(p_k=x)(M_q)\right)$ with $	M'(p_i)\geq 1$.
	
	Conversely, suppose that $p_i$ becomes marked in some reachable marking
	of the intervened PN.  Since $X_i$ is neither a root nor the intervened
	variable, the token in $p_i$ must have been produced by some transition
	$t_{i,c^{*}}$, where $c^{*}\in\min(\mathcal{C}_i^{+})$.
	For this transition to be enabled, every place corresponding to a
	component $c^{*}_\ell=1$ must have been simultaneously marked.
	By the induction hypothesis, these active parent states occur jointly
	in a parent configuration $c$ having positive support under the
	intervention.  This configuration necessarily satisfies
	$c^{*}\leq c$.
	Since $c^{*}\in\mathcal{C}_i^{+}$ and
	$\mathcal{C}_i^{+}$ is upward closed by the monotone support condition, it follows
	that $c\in\mathcal{C}_i^{+}$
	and therefore
	\[
	P\!\left(
	X_i=1\mid \operatorname{pa}(X_i)=c
	\right)>0.
	\]
	Since the parent configuration $c$ itself has positive interventional
	probability, the Markov factorization yields
	\[
	P\!\left(
	X_i=1\mid do(X_k=x),M_q
	\right)>0.
	\]
	
	The induction therefore establishes the equivalence for every variable
	in the mutilated graph.
\end{proof}

\begin{corollary}[Semantic equivalence]~\label{cor:seq}
For propositional (binary) variables under the Markov condition, every
\emph{positivity claim} of the form ``$X_j=1$ is possible under
$do(X_k=x)$'' has an exact counterpart in terms of marking reachability in
$T(\mathrm{BN})$.
\end{corollary}

The definition of $\mathrm{do\_PN}$ is
Pearl's mutilation written in PN vocabulary, and Theorem~\ref{thm:do} certifies that
the rewriting is faithful. In the BN the output of mutilation
is a distribution, while in the net it is a concrete marking of a concrete
individual, from which the counterfactual markings of
\S\ref{sec:translation} are obtained directly rather than by conditioning a
population-level object.

We can infer, from Corollary~\ref{cor:seq} that two BNs sharing
the same graph and the same pattern of zero/nonzero conditional
probabilities can differ in the exact values of those probabilities while
translating to the same $T(\mathrm{BN})$ with identical reachability sets. Since reachability is a qualitative notion that records whether
$P(X_j=1 \mid do(X_k=x))$ is positive or not, if we want to recover the magnitude of the intervention probability, we need an additional layer in our formalism. This  is precisely what the conformal layer
introduced in \S4 accomplishes.

\subsection{Local nets and the structural foundation of relevant controls}

\begin{proposition}[LocalNet and causal invariance]
\label{prop:localnet}
Let $q$ and $q'$ be two individuals with markings $M_q$ and $M_{q'}$ in
$T(\mathrm{BN})$. Then:

\begin{enumerate}[(a)]
  \item $\mathrm{LocalNet}(q) = \mathrm{LocalNet}(q')$ if and only if $M_q$
        and $M_{q'}$ generate identical sets of reachable places and fireable
        transitions.
  \item If $\mathrm{LocalNet}(q) = \mathrm{LocalNet}(q')$, then for any
        intervention $x$ and any outcome place $p_Y$:
        \[
          R\!\left(\mathrm{do\_PN}(p_T = x)(M_q)\right)(p_Y) =
          R\!\left(\mathrm{do\_PN}(p_T = x)(M_{q'})\right)(p_Y)
        \]
\end{enumerate}

\noindent where $R(\cdot)$ represents non-deterministic reachability. In that case, individuals with the same LocalNet have the same qualitative causal
effect structure under any intervention.
\end{proposition}

\begin{proof}[Proof]
(a) is immediate from Definition~\ref{def:localnet}. For (b): identical LocalNets imply
identical sets of fireable transitions. The PN intervention removes the same
arcs in both cases (arcs depend on network structure, not on marking). The
resulting initial markings differ only in $p_T$, and since all downstream
transitions are the same in both LocalNets, the reachable marking sets are
identical.
\end{proof}

\begin{proposition}[Structural foundation for conformal exchangeability]
\label{prop:exchangeability}
Let $\mathcal{R}_q^{*(t)}$ be the set of super-relevant controls of $q$ under
intervention $T=t$. Suppose that (i) for every
$i \in \mathcal{R}_q^{*(t)}$ it holds that
$\mathrm{LocalNet}(i) = \mathrm{LocalNet}(q)$, and (ii) the conditional
distribution of the outcome given the local mechanism is the same across
individuals sharing that LocalNet (\emph{local ergodicity}). Then the non-conformity scores
$\{V_i^{(t)}\}_{i \in \mathcal{R}_q^{*(t)}}$ are exchangeable with
$V_q^{(t)}$, guaranteeing conformal coverage:
\[
  \mathbb{P}\!\left(\tau_q \in \hat{C}(\tau_q)\right) \geq 1 - \alpha
\]
\end{proposition}

\begin{proof}[Proof]
Conformal coverage requires exchangeability of non-conformity scores within the
calibration set. Proposition~\ref{prop:localnet}(b) establishes that identical
LocalNets imply the same qualitative causal mechanism. By local ergodicity,
individuals governed by the same mechanism draw their outcomes from the same
conditional distribution, so the distribution of residuals
$|Y_i - \hat{\mu}^{(t)}(X_i)|$ is the same for
$i \in \mathcal{R}_q^{*(t)}$ and for $q$. Exchangeability follows, and the
standard conformal coverage guarantee \citep{vovk2005,barber2021} applies.
\end{proof}

LocalNet equality is a
purely structural condition, whereas exchangeability is a distributional
property that depends also on how the data are generated. What Proposition~\ref{prop:exchangeability} establishes is
the division of labor. The structural condition identifies \emph{which}
individuals are candidate controls, and the ergodicity condition is the
minimal probabilistic supplement under which that structural criterion yields
the coverage guarantee.

Proposition~\ref{prop:exchangeability} establishes
that \emph{if} controls share $q$'s LocalNet, \emph{then} coverage is
guaranteed. The empirical challenge of verifying or approximating LocalNet
equality from observational data is addressed by Proposition~\ref{prop:exchangeability}, which provides the formal criterion that those heuristics approximate, namely that two individuals
are valid controls for $q$ if and only if their LocalNets coincide.

\section{Discussion}
\label{sec:discussion}

\subsection{What the translation of BN to PN adds}

The main thesis defended in this paper is that PNs provide an operational, identity-preserving semantics in which individual causal histories and counterfactual executions are first-class objects, while these are not first-class objects in the standard probabilistic semantics of BN/SCM. SCMs already possess a semantics for individuals via their exogenous variables and potential outcomes. The translation into PNs makes operationally
explicit what the SCM leaves implicit, and that this explicitness has consequences. Four properties, none of which appears naturally in an SCM, carry the weight. 

First, \emph{individual history}: under ITP the token
reaching $p_Y$ carries the full record of the firings that produced it,
whereas the history of a unit is not an object of the SCM formalism at all.
Second, \emph{executable mechanisms}: a transition is something that fires,
so $q$'s counterfactual worlds are generated by running the net, not by
solving a system of equations whose individual-level form must first be
estimated. Third, \emph{concurrency and cycles}: both are native to PNs and
foreign to DAG-based models, as discussed below. Fourth, \emph{control
selection}: $\mathrm{LocalNet}(q)$ turns ``valid control for $q$'' into a
checkable structural criterion, the formal target of the heuristics of
\citet{delbianco2021,delbianco2023}.

\subsection{Relation to the structural account of actual causation}

The Halpern--Pearl structural model account of actual causation \citep{halpern2005}
might appear to already supply what we are calling for. It is evaluated
relative to a fixed context $u$, a complete assignment of the exogenous
variables, and is in this sense token-level: it answers a question about
\emph{this} instantiation of the model, not about a population. Two
differences separate our proposal from theirs. First, actual causation is a
qualitative, binary judgment, whether $X=x$ was \emph{a} cause of $Y=y$ in
context $u$, established via counterfactual dependence under contingencies.
It has no counterpart to the quantity we are after, $\tau_q = Y_q(1) -
Y_q(0)$. It does not ask how much $Y$ would have differed under the
alternative intervention, only whether $X$ made a difference at all. Second,
and more consequential for our purposes, the Halpern--Pearl semantics
presupposes that the full structural model, functional equations and context
$u$ included, is already given. It is a representational theory of what
causation \emph{is} once the individual model is fixed, not an epistemic or
estimation theory of how to \emph{obtain} that individual model from
population data. This is exactly the gap that opens under unbounded
heterogeneity, when $\text{Var}(Y \mid do(T=t), X=x) \to \infty$, population
regression cannot supply a reliable structural equation for $q$, and
refining the definition of actual causation does not by itself address this
estimation problem. The two-layer architecture developed here, PN structure
plus conformal quantification, is designed to fill precisely this gap, i.e. the
structural layer specifies which mechanisms are relevant to $q$
($\mathrm{LocalNet}(q)$) without assuming a parametric form, and the
quantitative layer supplies a finite-sample guaranteed estimate of the
counterfactual magnitude itself. The two accounts are consequently
complementary rather than competing. An actual-causation judgment could in
principle be evaluated on the marking $M_q$ once it has been estimated, but
the Halpern--Pearl framework does not itself solve the estimation problem
that motivates the present paper.

\subsection{Relation to Causal Kripke Models}

\citet{wang2025} and the present work share a starting point, namely that the Halpern--Pearl framework (\citep{halpern2005,pearl2009}) requires extension.
But the extensions go in complementary directions.

\citeauthor{wang2025} extend Halpern--Pearl \emph{upward}: they add modal operators and
possible-world semantics to reason about causation under epistemic uncertainty.
Their question is `what does an agent know about whether $X$ caused $Y$?' The
result is a modal logic of actual causality with sound and weakly complete
axiomatization. Their framework is type-level at its semantic base (it inherits Halpern--Pearl's causal models) but adds an epistemic layer for reasoning about alternative
possible worlds.

The present work extends Halpern--Pearl \emph{downward}, to the operational-token level
where causal mechanisms are concrete firings over concrete individuals. The
question is `what is the causal mechanism of individual $q$?' The result is a
process semantics (PN under ITP) that is token-level from its foundation.

The complementarity is the following. Causal Kripke Models are the appropriate
formalism when the problem is uncertainty about which causal model applies
(multiple possible worlds, partial agent knowledge). PNs under ITP are
appropriate when the causal model of $q$ is known or estimable, and the
question is the token-level effect of an intervention on that specific
individual. A unified theory combining both extensions, modal operators over
PN markings under ITP, is a natural direction for future work.

\subsection{Inverting Baetu's epistemic argument}

\citet{baetu2025} argues that knowledge of singular causation rests,
tacitly or explicitly, on prior contrasts between population-level patterns:
we cannot know that `Alice's smoking caused her cancer' without invoking
type-level regularities about smokers and cancer. Token causality is
epistemically dependent on type causality.

We accept this epistemic dependence while denying the ontological one. To \emph{estimate} $\tau_q$ we need a reference
population, the relevant controls $\mathcal{R}_q^{*(t)}$. In this sense,
knowledge of the ICE depends on population comparisons. But this epistemic
scaffolding does not constitute the causal fact itself. The PN under ITP makes
this distinction explicit. $M_q$ and its reachable markings are properties of
the individual (token-level ontology). $\mathcal{R}_q^*$ is the epistemic
scaffolding that permits estimation when the counterfactual cannot be directly
observed.

\subsection{Hall's two concepts and the PN formalism}

Hall's \citeyearpar{hall2004} dichotomy between \emph{dependence} (counterfactual,
token-level) and \emph{production} (mechanistic, process-level) is often
presented as a challenge for unified theories of causation. Any single formalism
seems to capture one at the expense of the other.

PNs under ITP integrate both. The dependence dimension is captured by the
difference in reachable markings: $R(M_q^{T=1})(p_Y) - R(M_q^{T=0})(p_Y)$ is
an overtly counterfactual quantity. The production dimension is captured by
token flow, a concrete sequence of transition firings, a causal
process that carries tokens from the intervention place to the outcome place.
Unlike Hall's dichotomy, the PN formalism does not force a choice since a single
run of the PN under ITP is simultaneously the mechanism of production (token
trajectory through transitions) and the ground of counterfactual dependence
(the reachable marking that defines what would have occurred under the
alternative intervention).

\subsection{Extensions and open problems}

\emph{Continuous variables.} The translation $T(\text{BN})$ assumes binary
variables. For continuous covariates $X \in \mathbb{R}^p$, the formalism extends
naturally via colored PNs (where tokens carry numerical values) or weighted
PNs (where arcs carry weights). Transitions become functions transforming values
rather than simply propagating tokens.  ITP persists in this case, only the implementation changes.

\emph{Non-monotone mechanisms.} When the conditions of Definition~\ref{def:msc} are not satisfied, non-minimal activation patterns of $X_i$ may not belong to $\mathcal{C}_i^{+}$ and thus, there might exist transitions that cannot be activated for certain configurations of $\mbox{pa}(X_i)$. This requires an alternative characterization of admissible transitions.

\emph{Cyclic causation.} PNs handle cycles naturally, unlike BNs, which are
acyclic by construction. For feedback systems, where the outcome of an
intervention retroacts on covariates, PNs offer a direct representation that
BN require extensions to achieve.

\emph{The ergodicity condition.} Proposition~\ref{prop:exchangeability} relies
on local ergodicity. Characterizing sufficient conditions for this assumption
in terms of $T(\text{BN})$ properties is an open problem connecting PN theory
with the transportability literature \citep{pearl2014} and invariant causal
prediction \citep{peters2016}.

\emph{Weakening LocalNet equality.} As noted in
\S\ref{sec:individualization}, exact equality of LocalNets could be relaxed to
isomorphism or to agreement within a bounded radius of the treatment and
outcome places. Determining which relaxations preserve
Proposition~\ref{prop:localnet}(b), and at what cost in the coverage guarantee
of Proposition~\ref{prop:exchangeability}, would widen the usable control sets
in applications.

\section{Conclusions}
\label{sec:conclusions}

This paper has argued for a precise thesis: when the object of causal inquiry
is the Individual Causal Effect $\tau_q = Y_q(1) - Y_q(0)$, the appropriate
formal framework is Petri Nets under the Individual Token Philosophy, not
Bayesian Networks under probabilistic semantics.  SCMs can define the ICE through realizations of their exogenous variables, and our correspondence theorems show that the two frameworks agree wherever both apply. Our central claim is that the PN formalism puts
the individual, its history, and its executable mechanisms in the foreground,
where the SCM framework reaches them only against the grain of its
distributional semantics.

The argument rests on a philosophical diagnosis and a technical correspondence.
The diagnosis (\S\ref{sec:typetoken}) is that Pearl's SCM framework contains a
token-level primitive, the $do(x)$ operator, but wraps it in distributional
expectations and returns to type-level conclusions. When treatment effects have
unbounded heterogeneity, the distributional layer collapses while the structural
layer remains intact. The ICE is the quantity that survives, a token-level,
counterfactual fact about a specific individual.

The technical correspondence (\S\ref{sec:translation}--\S\ref{sec:formal}) is
that PNs under ITP reproduce the full causal semantics of BNs, $d$-separation
(Theorem~\ref{thm:dsep}) and the $do(x)$ operator (Theorem~\ref{thm:do}), while
being natively token-level. The ITP/CTP distinction in PN theory (\S\ref{sec:typetoken})
is the formal counterpart of the type/token distinction in philosophy of
causation, and the collider case in the proof of Theorem~\ref{thm:dsep} is where this
distinction does formal work.

The individualization argument (\S\ref{sec:individualization}) shows that the
set of valid controls $\mathcal{R}_q^*$ for estimating $\tau_q$ is formally
characterized by equality of local subnets ($\mathrm{LocalNet}(q) =
\mathrm{LocalNet}(i)$). Proposition~\ref{prop:exchangeability} shows that this structural condition is exactly what is needed for the conformal coverage guarantee to hold.

The framework yields a principled two-layer account of individualized causality:
a structural layer (PN under ITP) that provides causal architecture without
distributional assumptions, and a quantitative layer (conformal prediction)
that provides finite-sample coverage guarantees without parametric assumptions.

Three directions stand out for future work. The most technically demanding is
to replace the local ergodicity assumption in Proposition~\ref{prop:exchangeability}
with verifiable conditions on $T(\text{BN})$. The most conceptually interesting
is to develop a unified framework combining the modal-epistemic extension of
\citet{wang2025} with the token-operational extension of the present work. The
most empirically immediate is to implement the $\mathrm{LocalNet}(q)$ criterion
as an algorithmic guide for control selection, replacing the heuristic similarity
measures currently used.


\section*{Declarations}

\noindent\textbf{Funding:} This work was supported by project PGI UNS
24-E172, Secretar\'ia General de Ciencia y T\'ecnica, Universidad Nacional
del Sur.

\noindent\textbf{Competing interests:} The authors have no competing
interests to declare that are relevant to the content of this article.

\noindent\textbf{Data, materials and/or code availability:} No datasets
were generated or analysed in the current study.

\noindent\textbf{Author contributions:} All authors contributed to the
conception and formal development of the paper, and read and approved the
final manuscript.

\bibliographystyle{plainnat}

\end{document}